%% file: egpaper_final.tex
\documentclass[10pt,twocolumn,letterpaper]{article}

\usepackage{cvpr}
\usepackage{times}
\usepackage{epsfig}
\usepackage{graphicx}
\usepackage{amsmath}
\usepackage{amssymb}
\usepackage{bbm}
\usepackage{booktabs}

\usepackage[breaklinks=true,bookmarks=false]{hyperref}

\cvprfinalcopy 

\def\cvprPaperID{****} 

\title{An attention-based multi-resolution model for prostate whole slide image classification and localization}

\author{Jiayun Li\\
University of California, Los Angeles\\
{\tt\small jiayunli@g.ucla.edu}
\and
Wenyuan Li\\
University of California, Los Angeles\\
{\tt\small liwenyuan.zju@gmail.com}
\and
Arkadiusz Gertych\\
Cedars-Sinai Medical Center\\
{\tt\small Arkadiusz.Gertych@cshs.org}
\and
Beatrice S. Knudsen\\
Cedars-Sinai Medical Center\\
{\tt\small Beatrice.Knudsen@cshs.org}
\and
William Speier\\
University of California, Los Angeles\\
{\tt\small speier@ucla.edu}
\and
Corey W. Arnold\\
University of California, Los Angeles\\
{\tt\small cwarnold@ucla.edu}
}
\begin{document}

\maketitle

\begin{abstract}
    \input{./tex/abstract}
\end{abstract}
\footnote{This paper appears at CVPR 2019 Towards Causal, Explainable and Universal Medical Visual Diagnosis (MVD) Workshop.}
\input{./tex/introduction}
\input{./tex/related_work}
\input{./tex/method}

\input{./tex/experiment.tex}
\input{./tex/result.tex}
\input{./tex/summary.tex}

\section{Acknowledgements}
The authors would like to acknowledge support from the UCLA Radiology Department Exploratory Research Grant Program (16-0003) and NIH/NCI awards R21CA220352 and P50CA092131. This research was also enabled in part by GPUs donated by NVIDIA Corporation.

{\small
\bibliographystyle{ieee_fullname}
\bibliography{egbib}
}

\end{document}

%% file: tex/abstract.tex
Histology review is often used as the `gold standard' for disease diagnosis. Computer aided diagnosis tools can potentially help improve current pathology workflows by reducing examination time and interobserver variability. Previous work in cancer grading has focused mainly on classifying pre-defined regions of interest (ROIs), or relied on large amounts of fine-grained labels. In this paper, we propose a two-stage attention-based multiple instance learning model for slide-level cancer grading and weakly-supervised ROI detection and demonstrate its use in prostate cancer. Compared with existing Gleason classification models, our model goes a step further by utilizing visualized saliency maps to select informative tiles for fine-grained grade classification. The model was primarily developed on a large-scale whole slide dataset consisting of 3,521 prostate biopsy slides with only slide-level labels from 718 patients. The model achieved state-of-the-art performance for prostate cancer grading with an accuracy of 85.11\% for classifying benign, low-grade (Gleason grade 3+3 or 3+4), and high-grade (Gleason grade 4+3 or higher) slides on an independent test set. 

%% file: tex/introduction.tex
\section{Introduction}
\label{sec:introduction}
Prostate cancer is the most common and second deadliest cancer in men in the U.S, accounting for nearly 1 in 5 new cancer diagnoses \cite{siegel2019cancer}. Gleason grading of biopsied tissue is a key component in patient management and treatment selection \cite{dall2012active, tosoian2016active}. The Gleason score (GS) is determined by the two most prevalent Gleason patterns in the tissue section. Gleason patterns range from 1 (G1), representing tissue that is close to normal glands, to 5 (G5), indicating more aggressive cancer. Patients with high risk cancer (\ie GS $> 7$ or G4 + G3) are usually treated with radiation, hormonal therapy, or radical prostatectomy, while those with low- to intermediate-risk prostate cancer (\ie GS $ < 6$ or G3 + G4) are candidates for active surveillance.

Currently, pathologists need to scan through a histology slide, searching for relevant regions on which to ascertain Gleason scores. This process can be time-consuming and prone to observer variability \cite{humphrey2004gleason, lavery2012gleason, huang2014gleason}. Therefore, computer aided diagnosis (CAD) tools can benefit clinical practice by identifying relevant regions and serving as a second reader. However, there are many unique challenges in developing CAD tools for whole slide images (WSIs), such as the very large image size, the heterogeneity of slide contents, the insufficiency of fine-grained labels, and possible artifacts caused by pen markers and stain variations.  

In this paper, we developed an attention-based multiple instance learning (MIL) model that can not only predict slide-level labels, but also provide visualization of relevant regions using inherent attention maps. Unlike previous work that relied on labor intensive labels, such as manually drawn regions of interest (ROIs) around glands, our model is trained using only slide-level labels, known as weak labels, which can be easily retrieved from pathology reports. In our proposed two-stage model, suspicious regions are detected at a lower resolution (\eg 5x), and further analyzed at a higher resolution (\eg 10x), which is similar to pathologists' diagnostic process. To the best of our knowledge, this is the first work that utilizes weakly-supervised attention maps and MIL to select ROIs and classify prostate biopsy slides. Our model was trained and validated on a dataset of 2,661 biopsy slides from 491 patients. The model achieved state-of-the-art performance, with a classification accuraccy of 85.11\% on a held-out testset consisting of 860 slides from 227 patients. 

%% file: tex/related_work.tex
\section{Related Work}
\label{related_work}
{\noindent \bf ROI-level classification.} Early work on WSI analysis mainly focused on classifying small ROIs, which usually were selected by pathologists from the large tissue slide \cite{farjam2007image, gorelick2013prostate, nguyen2011prostate}. However, this does not accurately reflect the true clinical task as to ensure completeness, pathologists must grade the entire tissue section rather than sub-selected representative ROIs. This makes models based on ROIs unsuitable for automated Gleason grading \cite{doyle2012boosted}.

{\noindent \bf Slide-level classification.} Instead of relying on ROIs, more recent research has focused on slide-level classification. Nagpal \etal \cite{nagpal2018development} developed a two-stage Gleason classification model. In the first-stage, a tile-level classifier was trained with over 112 million annotated tiles from prostatectomy slides. In the second stage, predictions from the first stage were summarized to a K-nearest neighbor classifier for Gleason scoring. They achieved an average accuracy of 70\% in four-class Gleason group classification (1, 2, 3, or 4-5). However, these methods \cite{doyle2012boosted, nagpal2018development, zhou2017large} required a well-trained tile-level classifier, which can only be developed on a dataset with manually drawn ROIs or slides with homogeneous tissue contents. Moreover, they did not incorporate information embedded in slide-level labels.   

To address these challenges, previous work has proposed using an MIL framework for WSI classification \cite{dietterich1997solving}, where the slide was represented as a bag and tiles within the bag were modeled as instances in the bag \cite{hou2016patch, campanella2018terabyte, ilse2018attention}. MIL models can be roughly divided into two types \cite{amores2013multiple, ilse2018attention}: instance-based \cite{ramon2000multi, maron1998framework, raykar2008bayesian} and bag-based \cite{andrews2003support,chen2006miles,dong2006comparison}. Bag-based methods project instance features into low-dimensional representations and often demonstrate superior performance for bag-level classification tasks \cite{ilse2018attention,amores2013multiple}. However, as bag-level methods lack the ability to predict instance-level labels, they are less interpretable and thus sub-optimal for problems where obtaining instance labels is important \cite{li2019knowledge,huang2019knowledge,schlemper2019attention,pesce2019learning}.   
Ilse \etal \cite{ilse2018attention} proposed an attention-based deep learning model that can achieve comparable performances to bag-level models without losing interpretability. A low-dimensional instance embedding, an attention mechanism for aggregating instance-level features, and a final bag-level classifier were all parameterized with a neural network. They applied the model on two histology datasets consisting of small tiles extracted from WSIs and demonstrated promising performance. However, they did not apply the model on larger and more heterogeneous WSIs. Also, attention maps were only used for a visualization method. Campanella \etal \cite{campanella2018terabyte} applied a instance-level MIL model for binary prostate biopsy slide classification (\ie cancer versus non-cancer). Their model was developed on a large dataset consisting of 12,160 biopsy slides, and achieved over 95\% area under the curve of the receiver operating characteristic (AUROC). Yet, they did not address the more difficult grading problem. Built upon the attention-based MIL model \cite{ilse2018attention}, our model further improves the attention mechanism with instance dropout \cite{singh2017hide}. Instead of only using the attention map for visualization, we utilize it to automatically localize informative areas, which then get analyzed at higher resolution for cancer grading. 

%% file: tex/method.tex
\section{Methods}
\label{sec:method}

Our model is trained with slide-level annotations in an MIL framework. Specifically, $k$ $N \times N$ tiles $x_i, i \in [1, k]$ can be extracted from a given WSI, which usually contains gigabytes of pixels. Different from supervised computer vision models, in which the label for each tile is provided, only the label for the slide (\ie the set of tiles) is available. This problem can be modeled with MIL by considering tiles as instances and the entire slide as a bag. In section \ref{subsec:method:mil}, we introduce the deep attention-based MIL model \cite{ilse2018attention} and the instance dropout method \cite{singh2017hide}. The attention-based informative tile selection method is discussed in section \ref{subsec:method:region}. The overview of our two-stage classification model is described in \ref{subsec:method:twostage}. 

\begin{figure*}[!tbh]
\begin{center}
\includegraphics[width=0.8\linewidth]{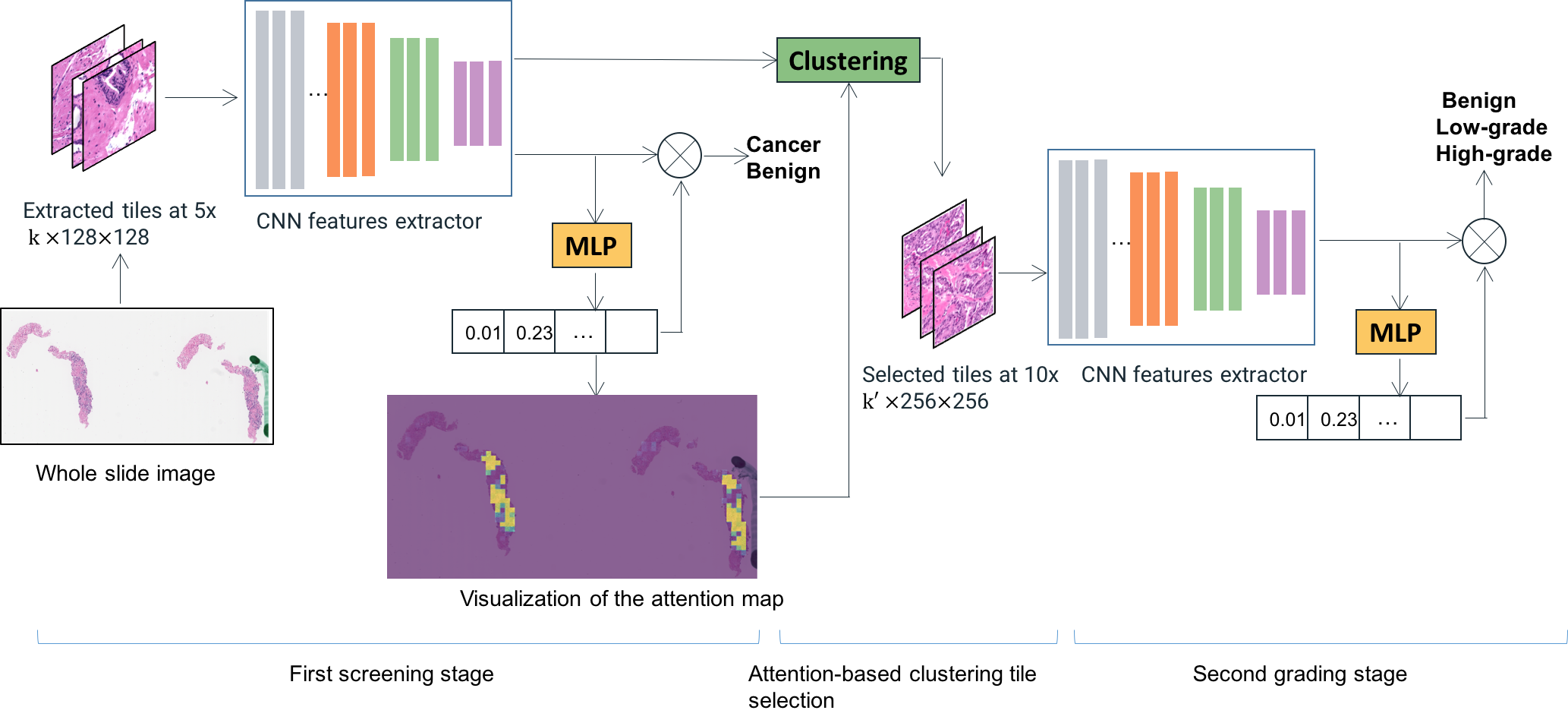}
\end{center}
\caption{The overview of our two-stage attention-based whole slide image classification model. The first stage is trained with tiles at 5x for cancer versus non-cancer classification. Informative tiles identified using instance features and attention maps from the first stage are selected to be analyzed in the second stage at a higher resolution for cancer grading.\label{fig:model_overview}}
\end{figure*}

\subsection{Attention-based multiple instance learning}
\label{subsec:method:mil}

The attention-based MIL model uses a convolutional neural network (CNN) as the backbone to extract instance-level features. An attention module $f(\cdot)$ is added before the softmax classifier to learn weight distribution $\boldsymbol{\alpha}= {\alpha_1, \alpha_2, ..., \alpha_k}$ for $k$ instances, which indicates importance of $k$ instances for predicting the current bag-level label $y$. The $f(\cdot)$ can be modeled by a multilayer perceptron (MLP). If we denote a set of $d$ dimensional feature vectors from $k$ instances as $\mathbf{V} \in \mathbb{R}^{k \times d}$, the attention for the $i$th instance can be defined in Eq(\ref{eq:attention}), 

\begin{align} \label{eq:attention}
\alpha_i = \textrm{Softmax}[\boldsymbol{U}^T(\mathrm{tanh}(\mathbf{W}_v\mathbf{v}_i^T))]
\end{align}
where $\boldsymbol{U} \in \mathbb{R}^{h \times n}$ and $\boldsymbol{W} \in \mathbb{R}^{h \times d}$ are learnable parameters, $n$ is the number of classes, and $h$ is the dimension of the hidden layer. Then each tile can have a corresponding attention value learned from the module. Bag-level embedding can be obtained by multiplying learned attentions with instance features. 

The attention distribution provides a way to localize informative tiles for the current model prediction. However, the attention-based MIL method suffers from the same problem as many saliency detection models \cite{zhang2018adversarial, singh2017hide,hou2018self,li2018tell}. Specifically, the model may only focus on the most discriminative input instead of all relevant regions. This problem may not have a large effect on the bag-level classification. Nevertheless, it could affect the integrity of the attention map and therefore affect the performance of our second stage model. To address this challenge, we utilize a similar method to \cite{singh2017hide}. During training, we randomly drop different instances in the bag by setting their pixel values to the mean RGB value of the training dataset \cite{singh2017hide}; in testing all instances will be used. This method forces the network to discover more relevant instances instead of only relying on the most discriminative ones. 

\subsection{Attention-based region selection}
\label{subsec:method:region}
In section \ref{subsec:method:mil}, we describe the attention-based MIL model and how to improve the learned attention map using instance dropout. One intuitive way to select informative tiles with attention maps is to rank them by attention values and select the top $k$ percentile. 
However, this method is highly reliant upon the quality of the learned attention maps, which may not be perfect, especially when there is no explicit supervision.   

To address this problem, we incorporate information from instance feature vectors $\mathbf{V}$. Specifically, instances are clustered into $n$ clusters based on instance features. Principle component analysis (PCA) is applied to reduce the dimension of features before clustering. Thus, instances that share similar semantic features will be grouped together. The average attention value for cluster $i$ with $m$ tiles can be computed $\bar{\alpha}_{i} = \frac{1}{m}\sum^{n}_{i=1}{\alpha}_i$ and normalized so that $\bar{\alpha}$ sums to 1. The intuition is that clusters with higher average attention are more likely to contain relevant information for slide classification (\eg given a cancerous slide, clusters containing stroma or benign glands should have lower attention values compared with those containing cancerous regions). Based on this, the number of tiles to be selected from each cluster can be determined by the total number of tiles and the average attention of the cluster.

\subsection{Two-stage whole slide image classification model}
\label{subsec:method:twostage}
In this section, we discuss how to incorporate the aforementioned methods into a two-stage WSI classification model. WSIs often contain several gigabytes of pixels, which practically impossible to fit into GPU memory. However, most regions on the WSIs are stroma or benign glands, which do not contribute to the final diagnosis. In clinical practice, pathologists usually scan through an entire slide at low magnification (\eg 5x), identify areas that may contain cancer, and closely examine these regions at a higher magnification (\eg 10x or 20x).   

Inspired by the workflow of pathologists, we developed a two-stage classification model. In the first screening stage, $128 \times 128$ tiles are extracted from each slide at 5x magnification and fed into a binary MIL model for cancer versus non-cancer classification. Informative tiles are identified by using attention maps and instance features from the 5x model as described in \ref{subsec:method:region}. Then, the second grading stage model uses selected tiles at 10x to classify the slide into benign, low-grade, or high-grade prostate cancer. Selected tiles are at the same location, but at a higher resolution as those in the first screening stage. Figure \ref{fig:model_overview} shows the overview of our two-stage WSI classification model.

%% file: tex/experiment.tex
\section{Experiment}
\label{sec:model}
In section \ref{subsec:exp:dataset}, we introduce the dataset and the preprocessing pipeline used. Details about model implementation and training are discussed in section \ref{subsec:exp:implementation}. 

\subsection{Dataset}
\label{subsec:exp:dataset}
{\noindent \bf Cedars Sinai dataset.} CNN feature extractors for both stages were pre-trained with a relatively small dataset with manually drawn ROIs from the Department of Pathology at Cedars-Sinai Medical Center (IRB approval numbers: Pro00029960 and Pro00048462) \cite{gertych2015machine, ing2018semantic, li2017multi, li2018path}. The dataset contains two parts.  
1) 513 tiles of size $1200 \times 1200$ extracted from prostatectomies of 40 patients, which contain low-grade pattern (Gleason grade 3), high-grade pattern (Gleason grade 4 and 5), benign (BN), and stromal areas. These tiles were annotated by pathologists at the pixel-level. 2) 30 WSIs from prostatectomies of 30 patients. These slides were annotated by a pathologist who circled and graded the major foci of tumor as either low-grade, high-grade, or BN areas.

The scanning objective for all slides and tiles was set at 20x (0.5 $\mu$m per pixel). To use this dataset for tile classification, we randomly sampled 11,595 tiles of size $256 \times 256$ at 10x from annotated regions. We will refer this dataset as the tile-level dataset in the following sections.     

{\noindent \bf UCLA dataset.} The MIL model is further trained with a large-scale dataset with only slide-level annotations. The dataset contains prostate biopsy slides from the Department of Pathology and Laboratory Medicine at the University of California, Los Angeles (UCLA). We randomly sampled a balanced number of low-grade, high-grade, and benign cases, resulting in 3,521 slides from 718 patients. We randomly divided the dataset based on patients for model training, validation, and testing to ensure the same patient would not be included in both training and testing. Labels for these slides were retrieved from pathology reports. For simplicity, we will refer this dataset as the slide-level dataset in the following sections.   

{\noindent \bf Data preprocessing.} Since WSIs may contain a lot of background regions and pen marker artifacts, we converted the slide at the lowest available magnification into HSV color space and thresholded on the hue channel to generate a mask for tissue areas. Morphological operations such as dilation and erosion were applied to fill in small holes and remove isolated points from tissue masks. Then, a set of instances (\ie tiles) for one bag (\ie slide) of size $256 \times 256$ at 10x was extracted from the grid with 12.5\% overlap. Tiles that contained less that 80\% tissue regions were removed from analysis. The number of tiles in the majority of slides ranged from 100 to 300. The same color normalization algorithm \cite{reinhard2001color} was performed on tiles from both UCLA and Cedars Sinai datasets. Tiles at 10x were downsampled to 5x for the first stage of model training.  
\subsection{Implementation Details}
\label{subsec:exp:implementation} 
{\noindent \bf Blue ratio selection.} Most previous work on WSI classification utilizes the blue ratio image to select relevant regions \cite{del2017convolutional, arvaniti2018coupling, lawson2019persistent}. The blue ratio image as defined in Eq(\ref{eq:blue_ratio}) reflects the concentration of the blue color, so it can detect regions with the most nuclei. 
\begin{align} \label{eq:blue_ratio}
\textrm{BR} = \frac{100 \times B}{1 + R + G} \times \frac{256}{1 + R + G + B}
\end{align}
where $R$, $G$, $B$ are the red, green and blue channels in the RGB image. The top $k$ percentile of tiles with highest blue ratio are selected. We used this method, br-two-stage, as the baseline for ROI detection.

{\noindent \bf CNN feature extractor.} As suggested by the previous study \cite{campanella2018terabyte}, we adopted the Vgg11 model with batch normalization (Vgg11bn) as the backbone for the feature extractor in both 5x and 10x models \cite{simonyan2014very}. The Vgg11bn was initialized with weights pretrained on ImageNet \cite{deng2009imagenet}. The feature extractor was first trained on the tile-level dataset for tile classification. After that, the fully connected layers were replaced by a $1 \times 1$ convolutional layer to reduce the feature map dimension, outputs of which were flattened and used as instance feature vectors $\mathbf{V}$ in the MIL model for slide classification. The batch size of the tile-level model was set to 50, the initial learning rate was set to $1e^{-5}$. Adam \cite{kingma2014adam} was used for model optimization. 

\begin{table*}[!tbh]
\begin{center}\caption{Model performances on whole slide image classification for prostate cancer\label{tab:compare_others}}
\begin{tabular}{@{} l *4c @{}}
\toprule
 \multicolumn{1}{c}{Models}  & Accuracy (\%) & Dataset & Classification task\\
 \midrule
 Zhou \etal \cite{zhou2017large}  & 75.00 & 368 slides  & G3 + G4 and G4 + G3 slides \\  
 Xu \etal \cite{xu2018automatic} & 79.00 & 312 slides  & GS 6, GS 7, and GS 8 slides \\  
 Nagpal \etal \cite{nagpal2018development} & 70.00 & 112 million patches and 1490 slides & 4 Gleason groups \\  
 \midrule
 Ours & 85.11 & 3521 slides & benign, low-grade, high-grade slides\\
 \bottomrule
\end{tabular}
\end{center} 
\end{table*}  

{\noindent \bf Two-stage classification model.} The first stage model was developed for cancer versus non-cancer classification. We transferred the knowledge from the tile-level dataset by initializing the feature extractor with learned weights. The feature extractor was initially fixed, while the attention module and classification layer were trained with a learning rate at $1e^{-4}$ for 10 epochs. Then we fine-tuned the last two convolutional blocks for the Vgg11bn model with a learning rate of $1e^{-5}$ for the feature extractor, and a learning rate of $1e^{-4}$ for the classifier for 90 epochs. Learning rates were reduced by 0.1 if the validation loss did not decrease for the last 10 epochs. The instance dropout rate was set to 0.5. Feature maps of size $512 \times 4 \times 4$ were reduced to $64 \times 4 \times 4$ after the $1 \times 1$ convolution, and then flattened to form a $1024 \times 1$ vector. A fully connected layer embedded it into a $1024 \times 1$ instance feature vector. The size of the hidden layer in the attention module $h$ was set to 512. The model with the highest accuracy on the validation set was utilized to generate attention maps. PCA was used to reduce the dimension of the instance feature vector to 32. K-means clustering was then performed to group similar tiles. The number of clusters was set to 4. Hyper-parameters were tuned on the validation set. Selected tiles at 10x were fed into the second-stage grading model. Similarly, we initialized the feature extractor with weights learned from the tile-level classification. The model was trained for five epochs with the feature extractor fixed. Other hyperparameters were the same as the first-stage model. 
Both tile- and slide-classification models were implemented in PyTorch 0.4, and trained using one NVIDIA Titan X GPU.

%% file: tex/result.tex
\section{Results}
\label{sec:result}  
\begin{figure}[!tbh]
\begin{center}
\includegraphics[width=0.6\linewidth]{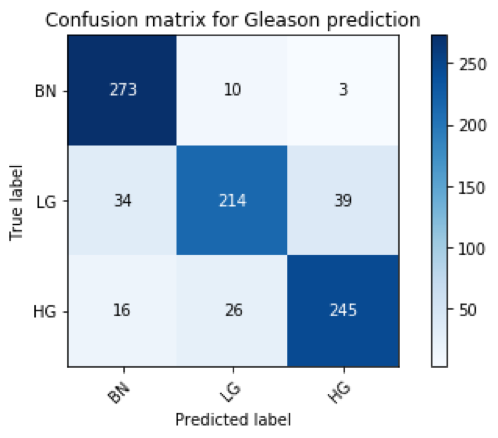}
\end{center}
\caption{Confusion matrix for Gleason grade classification on the test set\label{fig:model_cm}}
\end{figure}

We have summarized the performance of most state-of-the-art models for prostate WSIs classification in Table \ref{tab:compare_others}. The confusion matrix for our best model is shown in Figure \ref{fig:model_cm}. As shown in Table \ref{tab:compare_others}, the task of Zhou \etal's work \cite{zhou2017large} is the closet to the presented study, with the main difference being that we included a benign class. The work by Xu \etal can be considered relatively easy compared with our task, since differentiating G3 + G4 versus G3 + G4 is non-trivial \cite{nagpal2018development, zhou2017large} and often has the largest inter-observer variability. The model developed by Nagpal \etal \cite{nagpal2018development} achieved a lower accuracy compared with our model. However, their model predicted more classes, but relied on tile-level labels, which may not be directly comparable.   

We performed several experiments to evaluate the effects of different components on model performance. Specifically, in experiment \textit{att-two-stage}, we selected informative tiles based only on attention maps generated from the first stage model, while in the \textit{att-cluster-two-stage} model, both instance features and attention maps were used as discussed in section \ref{subsec:method:region}. Since blue ratio-based tile selection is the most commonly used method, we implemented the \textit{br-two-stage} model to evaluate the effectiveness of the attention-based ROI detection. To investigate the instance dropout, we trained another model without instance dropout, \textit{att-no-dropout}. To evaluate the contribution of knowledge transferred from the Cedars dataset, we trained a model without transfer learning. For simplicity, we denoted this model as \textit{no-transfer}. The \textit{one-stage} model was trained with tiles only from 5x.  

\begin{table}[!htb]
\begin{center}\caption{Test performances for different multiple instance learning models for whole slide image classification\label{tab:test_models}}
\begin{tabular}{@{} l *2c @{}}
\toprule
 \multicolumn{1}{c}{Models}  & Accuracy (\%) \\
 \midrule
one-stage  & 77.80 \\  
br-two-stage & 80.11 \\ 
 \midrule
att-two-stage & 81.86 \\ 
att-no-dropout & 79.65 \\
\midrule
no-transfer & 84.30 \\
att-cluster-two-stage & \bf{85.11} \\
 \bottomrule
\end{tabular}
\end{center}
\end{table}  

From Table \ref{tab:test_models}, we can see that the model with clustering-based attention achieved the best performance with the average accuracy over 7\% higher than the \textit{one-stage} model, over 5\% higher than the vanilla attention model (\ie \textit{att-no-dropout}). All two-stage models outperformed the \textit{one-stage}, which utilized all tiles at 5x to predict cancer grading. This is likely due to the fact that important visual features, such as those from nuclei, may only be available at higher resolution. As discussed in section \ref{subsec:method:mil}, attention maps learned in the weakly-supervised model are likely to be only focused on the most discriminative regions instead of the whole part, which could potentially harm model performance.  

\begin{figure*}[!tbh]
\begin{center}
\includegraphics[width=0.7\linewidth]{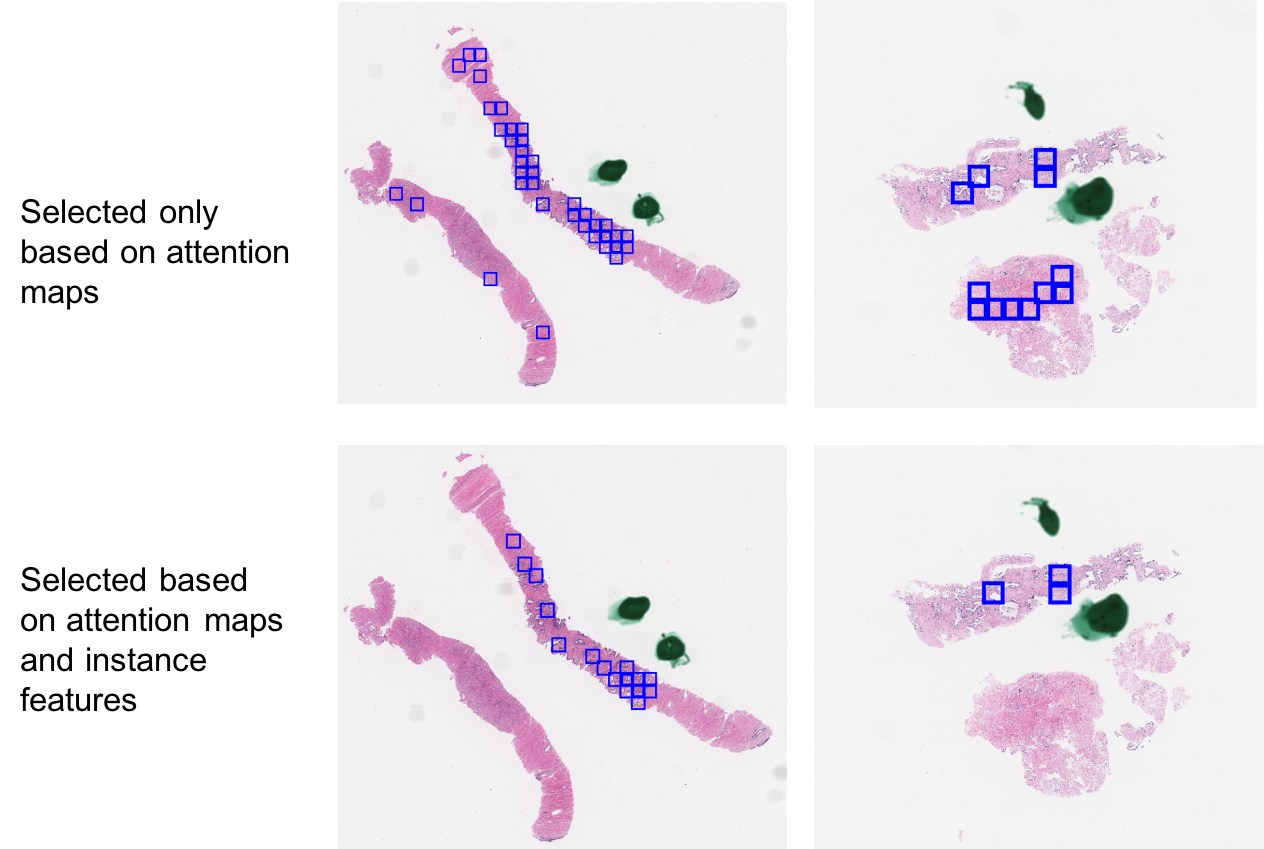}
\caption{Visualization of selected tiles based on different methods. Each blue box indicates one selected tile. \label{fig:att_cluster}}
\end{center}
\end{figure*}

\begin{figure*}[!tbh]
\begin{center}
\includegraphics[width=0.7\linewidth]{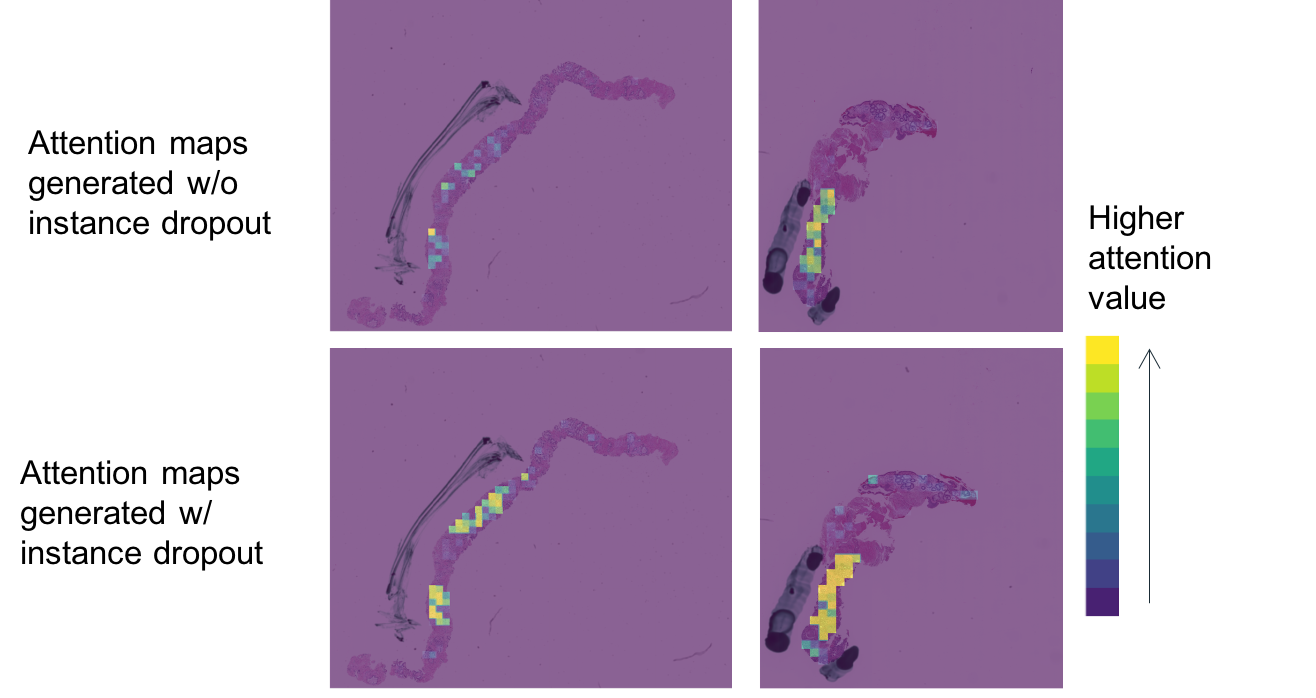}
\caption{Visualization of the model trained with or without instance dropout.\label{fig:att_dropout}}
\end{center}
\end{figure*}

As shown in Figure \ref{fig:att_cluster}, clustering with instance features reduced false positive tiles. Pen markers, which may indicate potential suspicious areas, were drawn by pathologists during the diagnosis. We did not use this information for model training, since it was not always available. In Figure \ref{fig:att_dropout}, we demonstrated the effect of instance dropout. The attention map trained without instance dropout failed to identify the entire region of interest. 

%% file: tex/summary.tex
\section{Discussion and Future Work}
\label{sec:summary}
In this paper, we developed an attention-based two-stage model for WSI classification on a large dataset with thousands of slides from hundreds of patients. Our model was trained to classify low-grade, high-grade, and benign slides. The model achieved an average accuracy of 85.11\%, which is over 5\% higher compared with the vanilla attention mechanism in \cite{ilse2018attention}, and we believe is state-of-the-art performance in prostate biopsy slide classification. In addition, the inherent attention mechanism enhances the interpretability of the classification results.   

There are some limitations of this work. Attention maps were only implicitly evaluated using the performance from the second stage model. Annotations or assessment from pathologists are needed for a better evaluation. Moreoever, we only included two resolutions (\ie 5x and 10x), which may not be sufficient to capture nucleoli-related features. In future work, higher resolutions will be used. As shown in the results, using transfer learning only slightly improved model performance. The reason could be that we only initialized the feature extractor with learned weights. However, more powerful transfer learning techniques such as \cite{ren2018adversarial} will be investigated in the future work.